\documentclass[fleqn]{cas-sc}

\usepackage[numbers]{natbib}
\usepackage{placeins}
\usepackage{verbatim}
\usepackage{caption}
\usepackage{subcaption}

\begin{document}
\let\WriteBookmarks\relax
\def\floatpagepagefraction{1}
\def\textpagefraction{.001}

\shorttitle{Prediction of OFCs via Ensemble Learning}

\shortauthors{J Zhang et~al.}

\title [mode = title]{Prediction of Oral Food Challenge Outcomes via Ensemble Learning}                      


%
\author[1]{Justin Zhang}






\author[2]{Deborah Lee}  

\author[3]{Kylie Jungles}    

\author[4]{Diane Shaltis}

\author[4,5,6,7,8]{Kayvan Najarian}

\author[2]{Rajan Ravikumar}

\author[2,3,9]{Georgiana Sanders}

\author[10]{Jonathan Gryak}[
    orcid=0000-0002-5125-7741]
\cormark[1]
\ead{jgryak@qc.cuny.edu} 

\affiliation[1]{organization={Department of Electrical and Computer Engineering, University of Michigan},
    city={Ann Arbor},
    state={MI},
    country={USA}}

\affiliation[2]{organization={Department of Internal Medicine, University of Michigan},
    city={Ann Arbor},
    state={MI},
    country={USA}}
\affiliation[3]{organization={Department of Pediatrics, University of Michigan},
    city={Ann Arbor},
    state={MI},
    country={USA}}
\affiliation[4]{organization={Department of Computational Medicine and Bioinformatics, University of Michigan},
    city={Ann Arbor},
    state={MI},
    country={USA}}

\affiliation[5]{organization={Michigan Institute for Data Science, University of Michigan},
    city={Ann Arbor},
    state={MI},
    country={USA}}
    
\affiliation[6]{organization={Department of Emergency Medicine, University of Michigan},
    city={Ann Arbor},
    state={MI},
    country={USA}}
\affiliation[7]{organization={Department of Computer Science and Engineering, University of Michigan},
    city={Ann Arbor},
    state={MI},
    country={USA}}

\affiliation[8]{organization={Max Harry Weil Institute for Critical Care Research and Innovation, University of Michigan},
    city={Ann Arbor},
    state={MI},
    country={USA}}

\affiliation[9]{organization={Mary H. Weiser Food Allergy Center, University of Michigan},
    city={Ann Arbor},
    state={MI},
    country={USA}}
    
\affiliation[10]{organization={Department of Computer Science, Queens College, City University of New York},
    city={New York},
    state={NY},
    country={USA}}

\cortext[cor1]{Corresponding author}



\begin{abstract}
Oral Food Challenges (OFCs) are essential to accurately diagnosing food allergy due to the limitations of existing clinical testing. However, some patients are hesitant to undergo OFCs, while those willing suffer from limited access to allergists in rural/community healthcare settings. Despite its success in predicting patient outcomes in other clinical settings, few applications of machine learning  to food allergy have been developed. Thus, in this study, we seek to leverage machine learning methodologies for OFC outcome prediction. Retrospective data was gathered from 1,112 patients who collectively underwent a total of 1,284 OFCs, and consisted of clinical factors including serum-specific Immunoglobulin E (IgE), total IgE, skin prick tests (SPTs), comorbidities, sex, and age.  Using these features, multiple machine learning models were constructed to predict OFC outcomes for three common allergens: peanut, egg, and milk. The best performing model for each allergen was an ensemble of random forest (egg) or Learning Using Concave and Convex Kernels (LUCCK) (peanut, milk) models, which achieved an Area under the Curve (AUC) of 0.91, 0.96, and 0.94, in predicting OFC outcomes for peanut, egg, and milk, respectively. Moreover, all such models had sensitivity and specificity values $\geq 89\%$. Model interpretation via SHapley Additive exPlanations (SHAP) indicates that specific IgE, along with wheal and flare values from SPTs, are highly predictive of OFC outcomes. The results of this analysis suggest that ensemble learning has the potential to predict OFC outcomes and reveal relevant clinical factors for further study.
\end{abstract}





\maketitle

\section{Introduction}

The prevalence of food allergies has increased in the United States over the past few decades. Food allergy affects 8\% of US children and 4\% of the general population \cite{boyce2010guidelines}. Because incorrect food allergy diagnosis has been shown to increase anxiety and unnecessary food restrictions for patients and their families \cite{akuete2017multicenter}, appropriate diagnosis and management of reactions to foods is important to reducing risk and improving quality of life.

The diagnostic evaluation of food allergy involves careful review of reaction history, physical examination, and allergy testing - including skin prick tests (SPTs) and the quantitation of specific food Immunoglobulin E (IgE). These allergy tests have high sensitivity and negative predictive value, but poor specificity and positive predictive value. The combined presence of multiple clinical risk factors for food allergy can improve the accuracy of diagnosis of true allergy \cite{sporik2000specificity}.

Given the limitations of these methods, oral food challenges (OFCs) remain the gold standard for determining the clinical response to food allergies in a monitored setting. However, the lack of widespread use of OFCs remains problematic. With approximately 4,000 board-certified allergists in the US, many patients with suspected food allergies have limited access to OFCs. While the majority of patients administered OFCs pass with minor or no symptoms, OFCs carry the risk of severe reactions including epinephrine-refractory anaphylaxis \cite{alviani2020anaphylaxis}, which requires them to be administered in a clinical setting. A recent study suggested that fear and anxiety among patients and caregivers are a significant barrier to considering OFCs \cite{hsu2020oral}. Furthermore, the risks associated with OFC may contribute to high no-show or cancellation rates, wasting limited clinical resources.

In many clinical domains machine learning has found great success in predicting patient outcomes \cite{belle2015big}. However, despite the challenges in OFC administration as discussed above, there have been few attempts to leverage machine learning for the development predictive food allergy models. In \cite{dunngalvin2011highly}, logistic regression models that incorporated the knowledge of prior OFC outcomes algo with common electronic health record data, including demographics, IgE levels, and SPTs, were shown to predict OFC outcomes with high accuracy. However, the reliance of prior OFC outcomes limits the use of these models on newly challenged patients. Another study \cite{pmid33326635} utilizing machine learning for the prediction of egg OFC outcomes achieved accuracy, sensitivity, and specificity values 72\%, 68\%, and 74\%, with such performance metrics likely due to the small data set of 61 patients. Thus, in this study, we seek to advance the application of machine learning to OFC outcome prediction for three common allergens - peanut, milk, and egg - by utilizing a large retrospective dataset combined with ensemble learning methods to maintain sensitivity while improving specificity. Such a model could be incorporated into a low-cost clinical decision support system, reducing the risks associated with administering OFCs, thereby increasing patient adoption and reducing the need for in-clinic OFC administration.

In light of the above objective, the primary contributions of this work are:
\begin{enumerate}
	\item A machine learning pipeline was created to identify clinical variables relevant to OFC prediction and to evaluate multiple machine learning models and cross-validation/ensemble strategies.
	\item The use of ensemble learning methods to achieve state-of-the-art performance in OFC prediction for three common food allergens - peanut, egg, and milk. These methods greatly improve specificity while maintaining sensitivity.
	\item The interpretation of the best performing machine learning models via SHAP has corroborated known clinical factors related to OFC prediction and identified a previously underutilized diagnostic - SPT flare size.
\end{enumerate}

The remainder of the paper is organized as follows. Section \ref{sec:related} provides an overview of the application of machine learning to allergy in general and prior work in food allergy in particular. Section \ref{sec:methods} details the dataset employed in this study, features selection methods, the machine learning algorithms utilized, and the construction of the cross-validation and ensemble models. The performance results of the trained predictive models are presented in Section \ref{sec:results}, wherein the best performing model for each allergen was an ensemble of random forest (egg) or LUCCK (peanut, milk), which achieved an Area under the Curve (AUC) for peanut, egg, and milk OFC prediction of 0.91, 0.96, and 0.94, respectively. The interpretation of these models, limitations on their applicability, and an analysis of model decision making via SHAP are discussed in Section \ref{sec:discussion}, including the significance of SPT flare size to OFC prediction. The results of this analysis suggest that ensemble learning has the potential to predict OFC outcomes with high performance and reveal relevant clinical factors for further study.
\section{Related Work}
\label{sec:related}
Precision medicine is an approach to managing medical conditions based on disease endotypes \cite{pmid28122115}. Defining relevant endotype biomarkers is an expanding area of research where machine learning has been employed. Machine learning applications have been used to find significant associations among large amounts of laboratory data and various clinical conditions, leading to advances in precision medicine \cite{rajkomar2019machine,qiu2019personalized,xiao2019comparison,pmid35452850,pmid35893293}. In allergy, machine learning applications have increased understanding of biomarkers and clinical correlations in asthma, atopic dermatitis, drug allergy, and food allergy \cite{pmid32702519,pmid33326635,pmid33721370,pmid35108754,pmid35640774,pmid35779061,pmid36267429}. Recent shifts in precision medicine have looked towards practical applications of machine learning algorithms to predict clinical outcomes at the point of care. Prior studies have used machine learning to predict outcomes in allergy challenges for aspirin and beta lactam allergies \cite{pmid32702519,pmid36267429}. Machine learning applications have also shown promise in predicting severe asthma and atopic dermatitis responses to biologic therapies \cite{pmid35640774,pmid35779061}. 

The first study in which machine learning was applied to food allergy outcomes \cite{pmid33326635} utilized retrospective data in heated egg challenges for egg allergy.  The training variables included age, sex, total serum IgE, egg proteins, serum-specific IgE levels, and outcomes of oral food challenges to heated egg in a small cohort of 67 children with egg allergy. Using extreme gradient boosting and support vector machine models, the authors reported sensitivity and specificity values of 0.51 to 0.68 and 0.66 to 0.74 respectively for predicting heated egg challenge outcomes, along with accuracies of up to 72\%. Those this work demonstrated the potential of applying machine learning in practical settings for food allergy, the generalizability of the study was limited by small sample size, few relevant laboratory variables, and minimal clinical data.

\section{Materials and Methods}
\label{sec:methods}

\subsection{Overview}
\begin{figure}[ht]
	\centering
		\includegraphics[scale=.4]{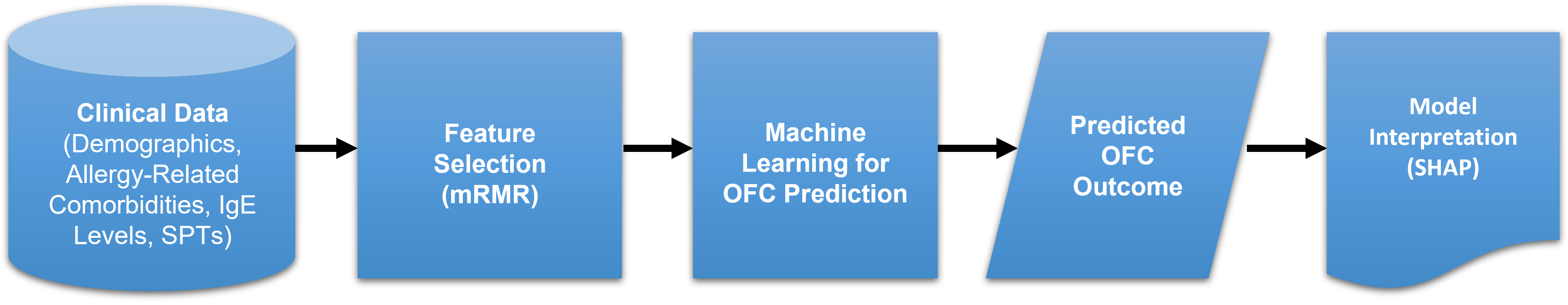}
	\caption{A schematic diagram of the proposed method.}
	\label{FIG:1}
\end{figure}

The overall framework of the proposed method is depicted in Figure \ref{FIG:1}. First, feature selection is performed to evaluate features by their redundancy, dependency, and relevance. Second, an optimal selection of features and hyperparameters is determined for the machine learning classifier to use in predicting the outcome of Oral Food Challenges (OFCs). The optimized ensemble models are then evaluated on the test set. The dataset is then shuffled and the entire process (feature selection, training and validation, testing) repeated 10 times, with an aggregate performance produced to compare with those of benchmark methods. Final model behavior is analyzed using SHapley Additive exPlanations (SHAP) values, which can be used to estimate the contribution of each variable to the final decision \cite{lundberg2017unified}. 

\FloatBarrier

\subsection{Data}

The secondary use of the data in this study was approved by the Michigan Medicine IRB under application HUM0019636. The patient cohort consists of patients between 0 and 80 years of age who underwent oral food challenges for evaluation of possible IgE-mediated allergies at the Michigan Medicine Allergy Clinic from January 2001 through March 2022.  Patients diagnosed with Food Protein-Induced Enterocolitis Syndrome (FPIES) were excluded. Food challenges for testing food intolerance or proximity were excluded, as were blinded challenges or those administered under food allergy immunotherapy trials. Physician labeling identified patients with passed, failed, and ambiguous/inconclusive OFC results. 

Due to changes in challenge protocol over time and idiosyncrasies between physicians, testing for IgE components is more prevalent and detailed for more recent OFCs. A portion of IgE and SPT data were read and entered into the system from stored paper records. Characteristics of these complexities in the older dataset include varying resolution (e.g., range and precision), and missing features (e.g., peanut components, wheal, and flare). The final set of serum-specific IgE and SPT factors utilized in this study for each food allergen are presented in Table \ref{specIgE}.

\begin{table}[width=.9\linewidth,cols=3,pos=h!]
	\caption{Serum-specific IgE test and Skin Prick Test (SPT) factors by food allergen. These common clinical test results are among the features utilized in constructing predictive models.}\label{specIgE}
	\begin{tabular*}{\tblwidth}{@{} LLL@{} }
		\toprule
		Allergen & IgE Test factors & SPT Factors\\
		\midrule
		Peanut & Total peanut & Peanut \\
		Egg & Egg White, Egg Yolk, Ovalbumin, Ovomucoid & Egg White (Needs Confirmation) \\
		Milk & $\alpha$-Lactalbumin, $\beta$-Lactoglobulin, Cow Milk, Casein, Whey & Cow Milk, Casein \\
		\bottomrule
	\end{tabular*}
\end{table}

OFCs of relevance to this study include those from peanut, combined cooked and baked egg (any egg), combined baked and other milk (any milk). The final database contained 495 peanut OFCs, 558 any egg OFCs, and 231 any milk OFCs. In total, 1,112 patients and 1284 OFCs were included in this study. There was a 86\% OFC pass rate and 62\% of patients were male and 38\% of patients were female. The median patient age was 5.6 years, and the lower and upper quartiles were 3.1 years and 5.6 years, respectively.

The available clinical variables in the dataset that were deemed salient to predicting OFCs were selected by board-certified allergists for inclusion in our analyses. These variables can be grouped into the following five categories: 
\begin{itemize}
	\item Demographics, including patient age, gender, race, and ethnicity
	\item Comorbidities
	\item Clinical rationale for administering a food challenge
	\item Total and serum-specific IgE tests
	\item Wheal and flare components of Skin Prick Tests (SPTs)
\end{itemize}
The number of variables per category and the total number per allergen are presented in Table \ref{categdata}, with peanut, egg, and milk challenges respectively having 41, 44, and 47 variables available for subsequent analysis.

\begin{table}[width=.9\linewidth,cols=7,pos=h!]
	\caption{The number of features per category by food allergen. The total corresponds to the number of features utilized in constructing predictive models for each allergen.}\label{categdata}
	\begin{tabular*}{\tblwidth}{@{} LLLLLLL@{} }
		\toprule
		Allergen & Demographics & Comorbidities & OFC Reason & IgE Tests & SPT Components & Total \\
		\midrule
		Peanut & 10 & 19 & 8 & 2 & 2 & 41 \\
		Egg & 10 & 19 & 8 & 5 & 2 & 44 \\ 
		Milk & 10 & 19 & 8 & 6 & 4 & 47 \\
		\bottomrule
	\end{tabular*}
\end{table}

In the following results, figures, and discussion, variables corresponding to comorbidities are suffixed with the phrase "Condition", while the clinical rationales for administering a food challenge are prefaced with the label "Reason:". An OFC is interpreted to have "passed" and belong to the \textit{positive class} when the patient did not experience a severe physiological reaction to the food allergen.

\FloatBarrier

\subsection{Data Preprocessing}
 OFCs that were missing feature values were removed from the dataset. SPT score, compiled from existing and recorded wheal, flare and pseudopod components, was deemed redundant and excluded from the data. All remaining challenges underwent data preprocessing, including the standardization of recorded serum-specific and total IgE values, such as `<0.35' being set to 0.175, and `>100' set to `101'. `Yes/No' and `Checked/Unchecked' features were converted to binary. Personally identifiable information (PII) such as dates and names were removed. Features such as `patient age’ were extrapolated from the patient birthday and OFC date features. The most recent test data prior to challenge were selected to ensure information is not result-informed, i.e., to prevent data leakage. The final dataset is then partitioned for feature selection, training/validation, and testing as depicted in Figure \ref{FIG:2a}.

\subsection{Feature Selection}
Features were ranked in importance using the minimum Redundancy Maximum Relevance (mRMR) method, which determine features that have maximum mutual information with respect to the given classes and minimum mutual information with each other \cite{peng2005feature}. This feature selection occurs prior to training and validation. The optimal feature combination is then found by surveying the model performance on the validation set with number of most important features through the elbow method heuristic. The entire process is repeated with every data shuffle.  

\subsection{Machine Learning}
Five machine learning algorithms were utilized to construct predictive models for OFCs for each food allergen. These are Naive Bayes, Logistic Regression, Support Vector Machines (SVM), Random Forest, and Learning Using Concave and Convex Kernels (LUCCK). Analyses were performed in Python and made use of various functionalities/implementations within the Scikit-learn framework \cite{sklearn_api}. The default implementations of the Gaussian Naive Bayes and $\ell_2$-penalized Logistic Regression models were utilized in this study. Additional details with respect to hyperparameter optimization strategies for the SVM, Random Forest, and LUCCK algorithms are presented in their respective subsections below.

\subsubsection{Support Vector Machine}
Support vector machines are a class of commonly used machine learning models that determine the maximum margin hyperplane separating the samples from distinct classes. The implementation of SVM utilized in this study was LIBSVM \cite{chang2011libsvm}. Both linear and radial basis function (RBF) kernels were evaluated, with the box constraint ($C$) and kernel width $\gamma$ (RBF kernel only) were chosen through twenty randomized searches over the hyperparameter space \cite{bergstra2012random}. Parameters were selected from logarithmically-uniform distributions for $C\in[10^{-3},10^{2}]$ and $\gamma\in[10^{-3},10^{2}]$.

\subsubsection{Random Forest}
Random forest is another commonly employed machine learning model that works well with high-dimensional data \cite{breiman2001random}. Five hyperparameters were optimized via grid search: number of trees, split criterion, minimum leaf size, number of variables to sample, and the maximum number of splits. Table \ref{RFhp} details the hyperparameter combinations evaluated during grid search optimization.

\begin{table}[width=.9\linewidth,cols=2,pos=h!]
\caption{Hyperparameters evaluated for the Random Forest models.}\label{RFhp}
\begin{tabular*}{\tblwidth}{@{} LL@{} }
\toprule
Parameter & Range\\
\midrule
Number of Trees & 20, 40, 80, 100, 160 \\
Split Criterion & Gini impurity, Information gain \\
Minimum Number of Leaf Node Observations & 1, 2, 5 \\
Number of Variables per Split & 6, 12, 24 \\
Maximal Number of Splits & Size of Training Set ($N$), $\log_2 N$ \\
\bottomrule
\end{tabular*}
\end{table}

\subsubsection{Learning Using Concave and Convex Kernels}
The Learning Using Concave and Convex Kernels (LUCCK) model \cite{sabeti2019learning} has been shown to outperform other machine learning methods when using features that may be prone to large measurement errors \cite{hernandez2021multimodal} or weakly correlated with desired outcomes \cite{alge2020predicting}. In the LUCCK algorithm, similarity (kernel) functions generate models for each feature separately. These functions take parameters that adjust their concavity and convexity within the feature space, which are then proportionally weighed according to information extracted from each feature.

More formally, suppose we have a training set $X\subseteq \mathbb{R}^n$ comprised of a disjoint union of labeled samples from $K$ classes: $X=C_1\cup C_2\cup\ldots\cup C_K$. A \textit{similarity function} $Q:\mathbb{R}^n\rightarrow\mathbb{R}$ measures the closeness of a vector $\mathbf{x}$ to the origin, and can be defined as a product $Q(\mathbf{x})=\prod_{i=1}^{n}{Q_i\left(x_i\right)}$, where $\mathbf{x}=\left(x_1,\ldots,x_n\right)\in\mathbb{R}^n$ and $Q_i\left(x_i\right)$ is a kernel function solely dependent upon $x_i$.  Following \cite{sabeti2019learning}, the kernel functions utilized in constructing LUCCK models for OFC prediction were of the form $Q_i\left(x_i\right)=\left(1+\lambda_ix^2\right)^{-\theta_i}$ with corresponding hyperparameters $\lambda_i,\theta_i>0$. The functions $Q_i\left(x\right)$ are fat-tailed distributions relative to a Gaussian, in that as $x\rightarrow\infty$, $Q_i\left(x\right)$ approaches zero at a slower rate than the function $e^{-\lambda_ix^2}$. The similarity function $Q(\mathbf{x})$ is then $$Q\left(\mathbf{x}\right)=\prod_{i=1}^{n}{\left(1+\lambda_ix^2\right)^{-\theta_i} }.$$
The parameters $\lambda_1,\lambda_2,\ldots,\lambda_n$ and $\theta_1,\theta_2,\ldots,\theta_n$ alter the convexity of the kernel functions, and can be chosen in a such a way as to optimize classification performance. Based on the chosen values the resulting kernel functions can be highly non-convex.

Once suitable values for the hyperparameters $\lambda_i$  and $\theta_i$ are chosen, the proximity of a feature vector $\mathbf{x}$ to a set $Y$ of training samples can be evaluated by aggregating the contributions of each element in $Y$: $$R\left(\mathbf{x},Y\right)=\sum_{y\in Y} Q\left(\mathbf{x}-\mathbf{y}\right).$$
A sample $\mathbf{x}$ is classified as class $k$, where $k$ is chosen such that $R\left(\mathbf{x},C_{k}\right)$ is maximal. Additional details on the LUCCK algorithm can be found in \cite{sabeti2019learning}.

In the proposed framework, the hyperparameters $\lambda_i$  and $\theta_i$ controlling the convexity of the kernel functions were chosen through twenty randomized searches over the hyperparameter space \cite{bergstra2012random}. Parameters were selected from logarithmically-uniform distributions for $\theta_i\in[10^{-2},10^{3}]$ and $\lambda_i\in[10^{-2},10^{4}]$.

\subsubsection{Model Training and Validation}
The experimental design follows a 3-fold patient-wise cross-validation scheme. To correct the 86\%/14\% positive/negative class imbalance, the training data was enriched to class parity using the SMOTE algorithm, which synthesizes new samples for the minority class from currently existing samples \cite{chawla2002smote}. No resampling was performed on the feature selection or test sets. For models sensitive to feature scaling (LR, SVM, and LUCCK), the training data was standardized, after which the means and standard deviations calculated on the training set were used to standardize the validation and test sets. For models employing hyperparameter optimization (SVM, Random Forest, and LUCCK), the hyperparameters resulting in the greatest AUC across all three folds were chosen to construct the final cross-validation model, which utilized all three folds of the training/validation data.

\begin{figure}[ht]
	\centering
    \begin{subfigure}{.45\textwidth}
        \centering
    		\includegraphics[scale=.45]{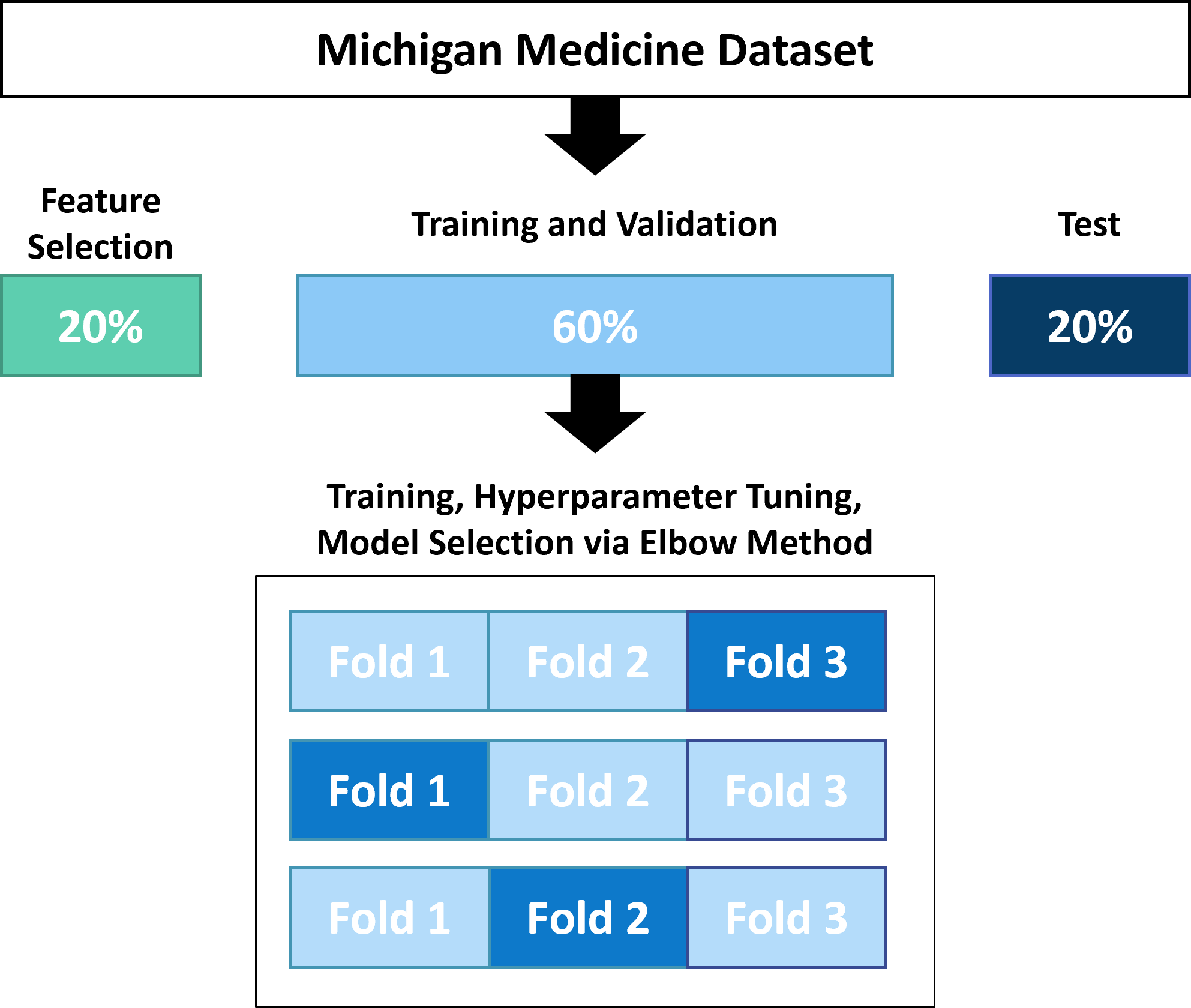}
        \caption{}
    	\label{FIG:2a}    
    \end{subfigure}
    \hfill
    \begin{subfigure}{.45\textwidth}
        \centering
     	\includegraphics[scale=.45]{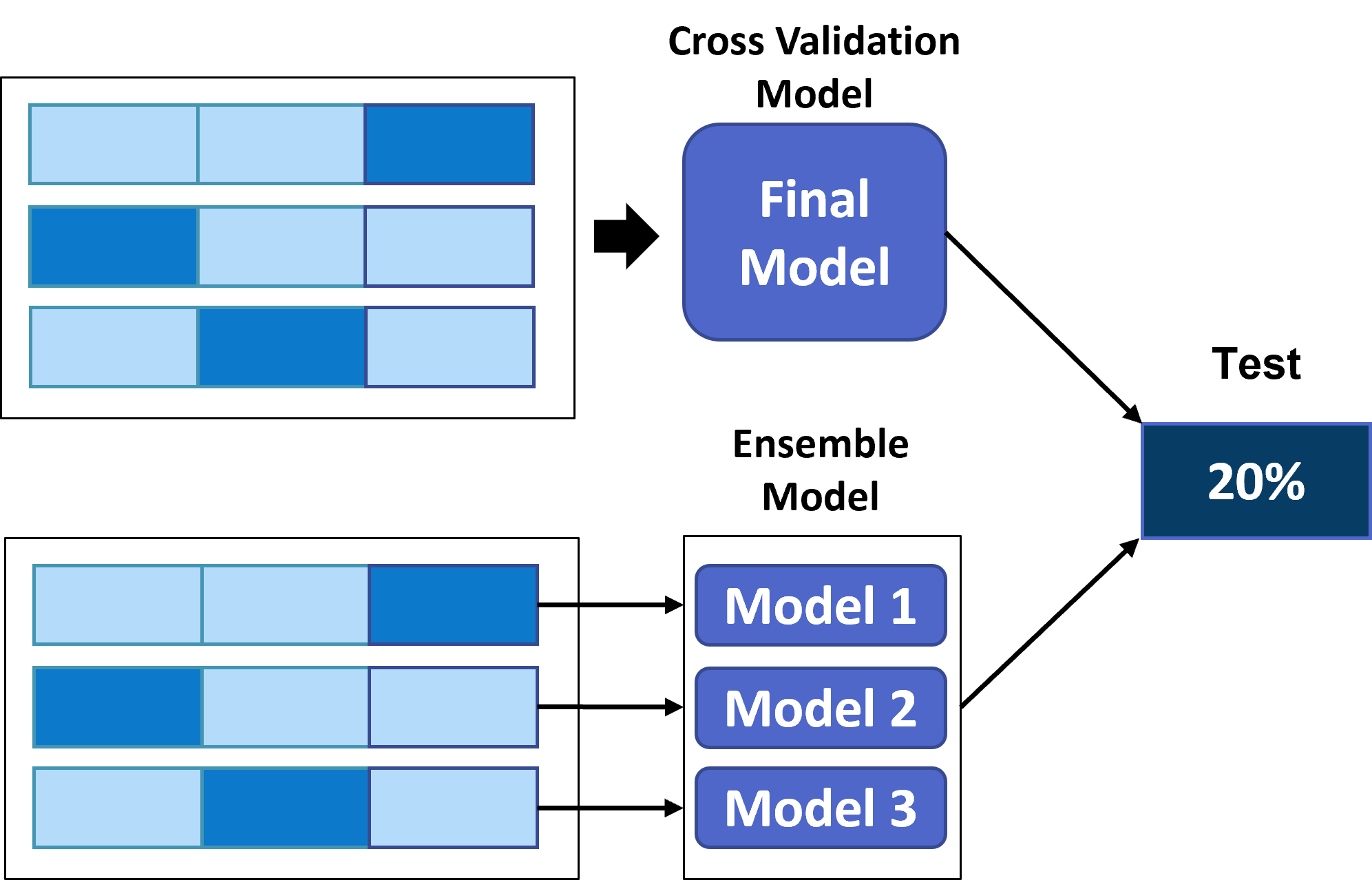}
    	\caption{}
    	\label{FIG:2b}   
    \end{subfigure}
    \caption{(a) The partitioning of the dataset; (b) the construction of the cross-validation and ensemble models.}
\end{figure}

Additionally, the cross-validation method was compared with an ensemble method, wherein the three models constructed during cross-validation are combined to form a single ensemble model, with the ensemble's final prediction achieved via \textit{majority voting}, i.e., the mode of their test set predictions. The model creation process is depicted in Figure \ref{FIG:2b}.

Additional model construction methods were also considered. In one set of experiments, OFCs were stratified by age and test date values (for SPT, total and specific serum IgE tests) to assess their influence on model performance. Trials were conducted where models were trained and tested that including OFC data from patients between the ages of 0 to 2 years, 2 to 18, and above 18 years, in an effort to address distributional drift due to underlying immunologic changes. Creating separate models for these age brackets did not improve model performance. Other variations, such as the inclusion of the time elapsed between the collection of IgE/SPT data and the OFC, nor adding an indicator variable marking tests occurring more than 6 months prior to the OFC, had a significant impact on model performance. As such, these models are not given further consideration in this work.

\FloatBarrier

\subsection{Model Interpretation}
SHapley Additive exPlanations (SHAP) is a popular post-hoc model interpretation technique \cite{lundberg2017unified}. SHAP utilizes cooperative game theory to distribute predicted outcomes through the calculation of Shapley values for each feature. Positive Shapley values indicate greater contributions towards a predicted outcome (i.e., the positive class), and vice versa for negative values. As such, SHAP analysis can provide insight into how individual features contribute to a model's decision making process. 
In this study, SHAP values were calculated using Python implementation of the kernelSHAP method as introduced in  \cite{lundberg2017unified}. Each feature's total contribution to a model's prediction for each sample in the test set is then summed and averaged to determine which features are most influential in the prediction of OFC outcomes.

\section{Results}
\label{sec:results}
\begin{figure}[ht]
	\centering
	\includegraphics[scale=.4]{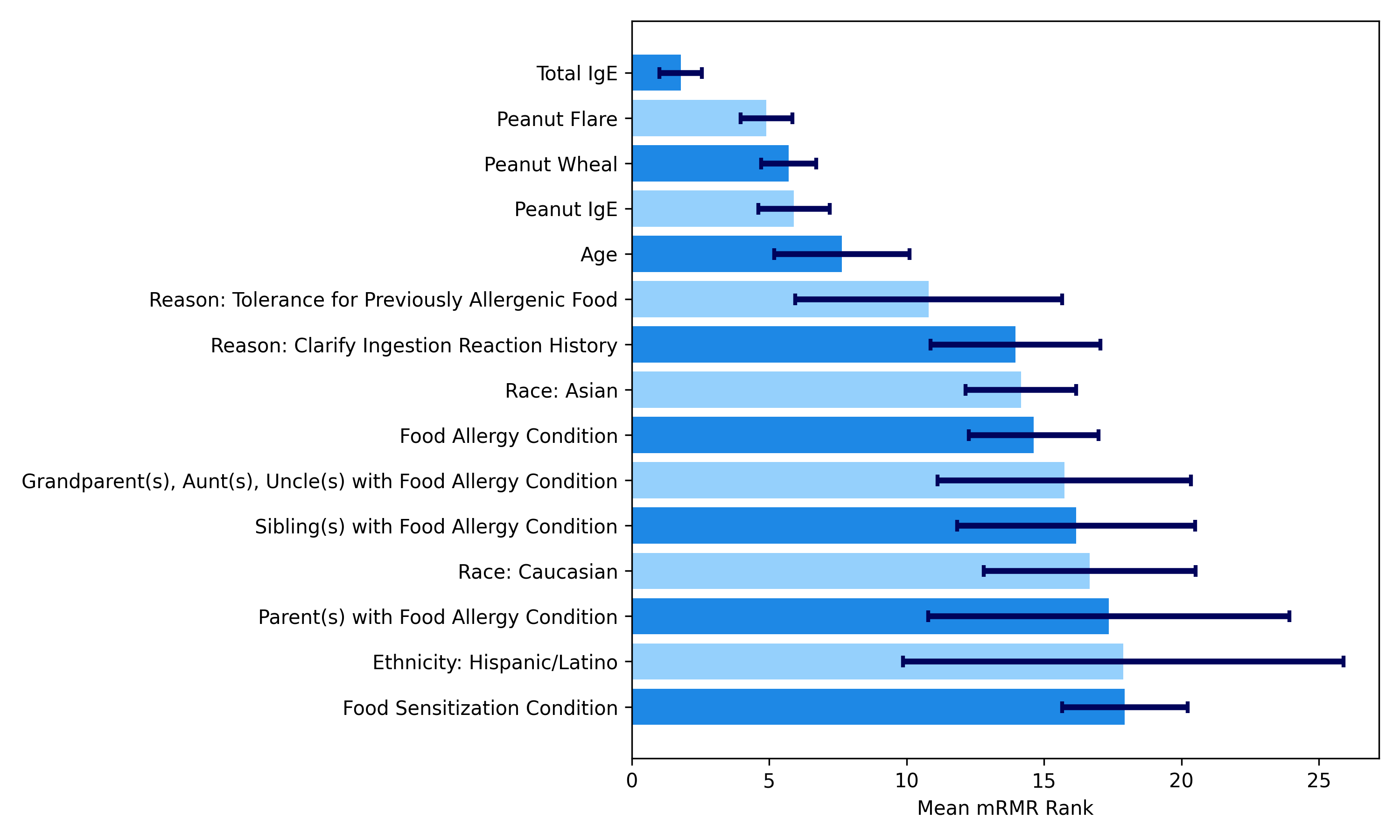}
	\caption{Mean mRMR feature rankings for peanut OFCs. Lower rankings are better.}
	\label{FIG:3}
\end{figure}

\begin{figure}[ht]
	\centering
	\includegraphics[scale=.4]{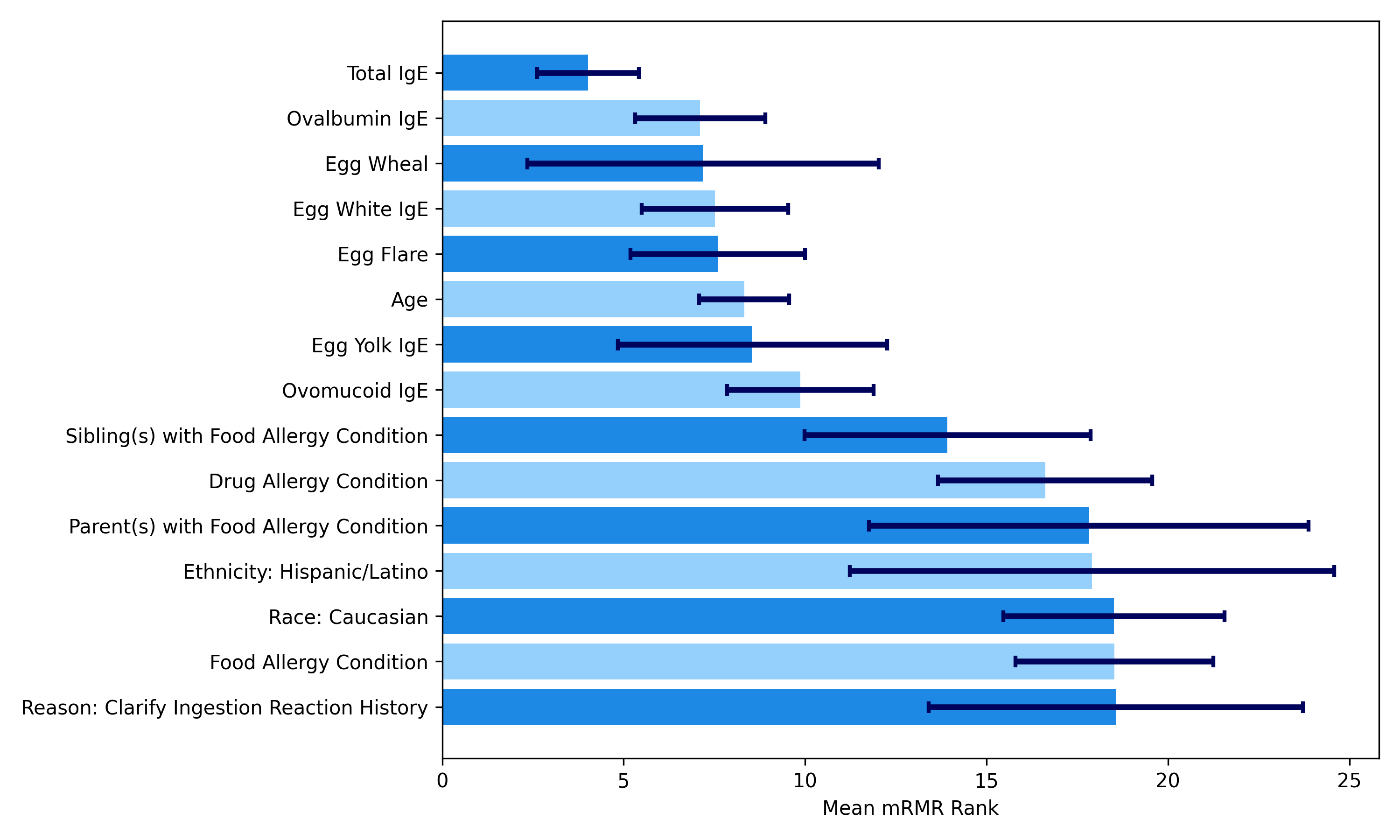}
	\caption{Mean mRMR feature rankings for egg OFCs. Lower rankings are better.}
	\label{FIG:4}
\end{figure}

\begin{figure}[ht]
	\centering
	\includegraphics[scale=.4]{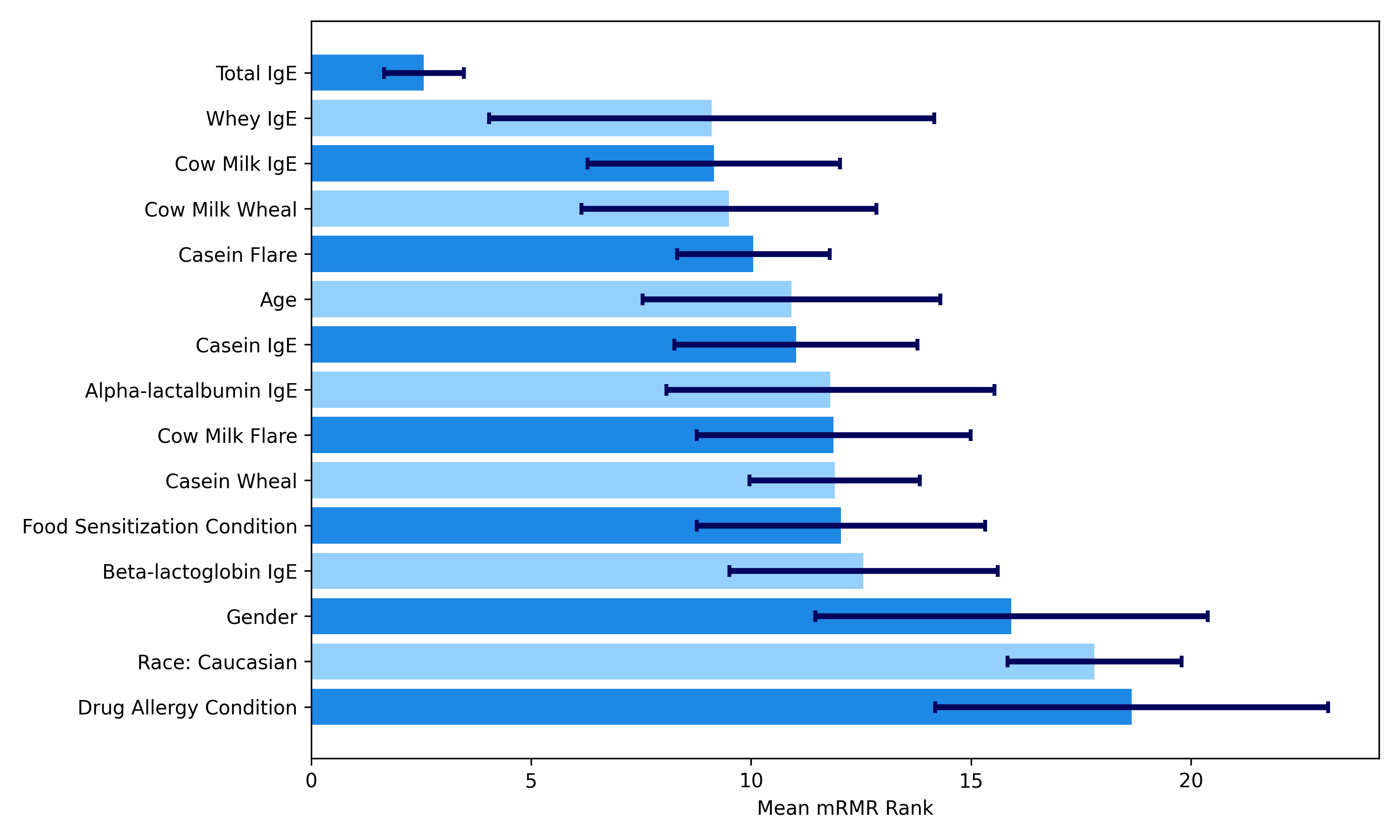}
	\caption{Mean mRMR feature rankings for milk OFCs. Lower rankings are better.}
	\label{FIG:5}
\end{figure}

The mean feature importance rankings as determined by mRMR prior to model creation, for peanut, egg, and milk OFCs, are depicted in Figures \ref{FIG:3},\ref{FIG:4}, and \ref{FIG:5}. In these figures, lower mRMR ranking indicates greater feature importance, with the error bars on each feature corresponding to the standard deviation across the ten permutations of the dataset. For all three food allergens, Total IgE was determined by mRMR to be the most significant feature.

Additional top features for peanut OFCs include SPT wheal and flare measurements, along with peanut specific IgE and patient age. Similar patterns occurred for egg and milk OFCs. Ovalbumin and egg white specific IgEs along with SPT results were the most relevant for egg OFCs, while whey and cow milk specific IgEs, along with SPT results, were most relevant for milk OFCs. In evaluating the performance of models constructed using varying subsets of features, the top fifteen (15) features as ranked by mRMR were found to yield the best performance on the validation set via the elbow heuristic. As such, all subsequent models were created using the top 15 features for each respective food allergen.

Performance metrics, including Area under the Curve (AUC) and Positive Predictive Value (PPV), of the five evaluated algorithms for predicting OFC outcomes for all three food allergens are reported in Tables \ref{tab:cv} and \ref{tab:ensemble}. Table \ref{tab:cv} displays the performance of models trained and tested using the cross-validation method, while Table \ref{tab:ensemble} displays the performance of models trained and tested with the ensemble method.

\begin{table}[ht]
	\caption{A comparison of mean (standard deviation) performance metrics of the \textit{cross-validation models} in predicting outcomes for peanut, milk, and egg OFCs. \textbf{Bolded} metric values indicate the highest performance achieved.}
	\label{tab:cv}
\begin{tabular}{lllllll}
	\toprule
	{} &          AUC &           F1 &     Accuracy &  Sensitivity &  Specificity &          PPV \\
	\midrule
	Naive Bayes         &  0.61 (0.05) &  0.71 (0.02) &  0.59 (0.03) &  0.58 (0.03) &  \textbf{0.63} (0.10) &   0.90 (0.02) \\
	Logistic Regression &  0.67 (0.07) &  0.86 (0.02) &  0.78 (0.03) &  0.83 (0.03) &  0.52 (0.13) &   0.90 (0.02) \\
	SVM                 &  0.65 (0.08) &  0.87 (0.01) &  0.78 (0.02) &  0.84 (0.04) &  0.45 (0.18) &  0.89 (0.03) \\
	Random Forest       &  \textbf{0.67} (0.04) &  \textbf{0.88 }(0.02) &  \textbf{0.81} (0.04) &  \textbf{0.87} (0.04) &  0.47 (0.09) &   \textbf{0.90} (0.01) \\
	LUCCK               &  0.66 (0.08) &  0.84 (0.03) &  0.75 (0.05) &  0.79 (0.05) &  0.53 (0.14) &   0.90 (0.03) \\
	\bottomrule
	\multicolumn{7}{c}{a) Peanut OFCs}
\end{tabular}
\begin{tabular}{lllllll}
	\toprule
	{} &          AUC &           F1 &     Accuracy &  Sensitivity &  Specificity &          PPV \\
	\midrule
	Naive Bayes         &  0.57 (0.06) &  0.73 (0.02) &  0.61 (0.03) &  0.62 (0.03) &  0.52 (0.11) &  0.91 (0.02) \\
	Logistic Regression &  0.62 (0.06) &  0.85 (0.04) &  0.76 (0.05) &  0.80 (0.06) &  0.45 (0.10) &  0.92 (0.02) \\
	SVM                 &  0.60 (0.05) &  0.90 (0.02) &  0.83 (0.03) &  0.90 (0.04) &  0.29 (0.12) &  0.90 (0.01) \\
	Random Forest       &  \textbf{0.68} (0.05) &  \textbf{0.92} (0.02) &  \textbf{0.86} (0.03) & \textbf{0.91} (0.04) &  0.45 (0.12) &  \textbf{0.93} (0.01) \\
	LUCCK               &  0.65 (0.05) &  0.85 (0.02) &  0.75 (0.03) &  0.78 (0.04) &  \textbf{0.52} (0.10) &  0.92 (0.01) \\
	\bottomrule
	\multicolumn{7}{c}{b) Egg OFCs}
\end{tabular}

\begin{tabular}{lllllll}
	\toprule
	{} &          AUC &           F1 &     Accuracy &  Sensitivity &  Specificity &          PPV \\
	\midrule
	Naive Bayes         &  0.55 (0.09) &  0.62 (0.09) &  0.50 (0.09) &  0.48 (0.11) &  0.62 (0.17) &  0.89 (0.05) \\
	Logistic Regression &  0.62 (0.07) &  0.86 (0.03) &  0.78 (0.05) &  0.83 (0.07) &  0.42 (0.19) &  0.91 (0.02) \\
	SVM                 &  0.56 (0.09) &  0.89 (0.03) &  0.81 (0.05) &  0.89 (0.06) &   0.22 (0.20) &  0.88 (0.02) \\
	Random Forest       &  \textbf{0.64} (0.07) &  \textbf{0.91} (0.02) &  \textbf{0.84} (0.04) &  \textbf{0.91} (0.06) &  \textbf{0.37} (0.18) &  \textbf{0.91} (0.02) \\
	LUCCK               &  0.61 (0.08) &  0.88 (0.04) &  0.79 (0.05) &  0.86 (0.08) &  0.37 (0.19) &   0.90 (0.02) \\
	\bottomrule
	\multicolumn{7}{c}{c) Milk OFCs}
\end{tabular}
\end{table}
As show in Table \ref{tab:cv}, random forest achieved the best AUC for each food allergen using the cross-validation method, yielding mean AUCs (standard deviations) of 0.67 (0.04), 0.68 (0.05), and 0.64 (0.07) for peanut, egg, and milk OFCs respectively. Both logistic regression and LUCCK had similar performances to random forest, while SVM performing slightly worse and Naive Bayes having the lowest performance in terms of AUC. All models had high PPV (precision) of at least 0.90 with small standard deviations (0.01-0.03). Most models had low specificity, indicating their inability to correctly predict OFC failures, with the highest specificity being achieved by the Naive Bayes model for peanut OFCs, while SVM having the lowest specificity of 0.29 for egg OFCs.

\begin{table}[ht]
		\caption{Mean (standard deviation) performance metrics of the \textit{ensemble models} in predicting outcomes for peanut, milk, and egg OFCs. \textbf{Bolded} metric values indicate the highest performance achieved.}
	\label{tab:ensemble}
\begin{tabular}{lllllll}
	\toprule
	{} &          AUC &           F1 &     Accuracy &  Sensitivity &  Specificity &          PPV \\
	\midrule
	Naive Bayes         &  0.64 (0.04) &  0.72 (0.03) &  0.62 (0.03) &  0.60 (0.03) &  0.69 (0.08) &  0.91 (0.02) \\
	Logistic Regression &  0.78 (0.06) &  0.90 (0.02) &  0.83 (0.03) &  0.86 (0.03) &  0.71 (0.12) &  0.94 (0.02) \\
	SVM                 &  0.84 (0.04) &  0.92 (0.01) &  0.87 (0.02) &  0.89 (0.03) &  0.78 (0.08) &  0.96 (0.02) \\
	Random Forest       &  0.90 (0.05) &  \textbf{0.96} (0.01) &  \textbf{0.93} (0.02) &  \textbf{0.95} (0.01) &  0.85 (0.09) &  0.97 (0.02) \\
	LUCCK               &  \textbf{0.91} (0.04) &  0.94 (0.02) &  0.90 (0.03) &  0.89 (0.03) &  \textbf{0.92} (0.07) &  \textbf{0.98} (0.01) \\
	\bottomrule
	\multicolumn{7}{c}{a) Peanut OFCs}
\end{tabular}

\begin{tabular}{lllllll}
	\toprule
	{} &          AUC &           F1 &     Accuracy &  Sensitivity &  Specificity &          PPV \\
	\midrule
	Naive Bayes         &  0.60 (0.06) &  0.74 (0.03) &  0.61 (0.04) &  0.62 (0.04) &  0.58 (0.08) &  0.92 (0.02) \\
	Logistic Regression &  0.70 (0.05) &  0.87 (0.03) &  0.78 (0.04) &  0.81 (0.04) &  0.60 (0.10) &  0.94 (0.01) \\
	SVM                 &  0.76 (0.06) &  0.95 (0.01) &  0.92 (0.02) &  0.97 (0.02) &  0.55 (0.12) &  0.94 (0.02) \\
	Random Forest       &  \textbf{0.96} (0.02) &  0.99 (0.01) &  \textbf{0.98} (0.01) &  \textbf{0.98} (0.01) &  \textbf{0.95} (0.04) &   \textbf{0.99} (0.00) \\
	LUCCK               &  0.92 (0.03) &  \textbf{0.94} (0.02) &  0.89 (0.04) &  0.89 (0.04) &  0.95 (0.05) &  0.99 (0.01) \\
	\bottomrule
	\multicolumn{7}{c}{b) Egg OFCs}
\end{tabular}

\begin{tabular}{lllllll}
	\toprule
	{} &          AUC &           F1 &     Accuracy &  Sensitivity &  Specificity &          PPV \\
	\midrule
	Naive Bayes         &  0.61 (0.05) &  0.63 (0.03) &  0.51 (0.03) &  0.48 (0.05) &  0.73 (0.13) &  0.93 (0.03) \\
	Logistic Regression &  0.73 (0.08) &  0.88 (0.04) &  0.81 (0.05) &  0.84 (0.07) &  0.63 (0.19) &  0.94 (0.03) \\
	SVM                 &  0.79 (0.08) &  0.96 (0.01) &  0.92 (0.02) &  0.97 (0.02) &  0.60 (0.15) &  0.94 (0.02) \\
	Random Forest       &  0.85 (0.10) &  \textbf{0.97} (0.02) &  \textbf{0.95} (0.03) &  \textbf{0.98} (0.02) &   0.72 (0.20) &  0.96 (0.03) \\
	LUCCK               &  \textbf{0.94} (0.03) &  0.96 (0.03) &  0.94 (0.05) &  0.94 (0.06) &  \textbf{0.95} (0.08) &  \textbf{0.99} (0.01) \\
	\bottomrule
	\multicolumn{7}{c}{c) Milk OFCs}
\end{tabular}
\end{table}
The performance metrics for the ensemble models are presented in Table \ref{tab:ensemble}. The ensemble models had better performance for all machine learning methods and all food allergens than those created using the cross-validation method. However, the amount of improvement varied significantly by algorithm and metric. Naive Bayes saw the least improvement among all models in terms of AUC, with 3\% improvement for the peanut and egg OFC models and a 7\% improvement in milk OFC prediction. However, all models had significant improvements in specificity in their ensemble versions, with Naive Bayes, logistic regression, SVM, Random Forest, and LUCCK respectively obtaining 8\%, 18\%, 32\%, 41\% and 47\% increases on average across all food allergens.

The best performing ensemble models, and the \textit{best overall}, for all allergens were the random forest and LUCCK ensembles. In terms of AUC, the LUCCK ensemble models achieved the highest performance in OFC prediction for peanut - 0.91 (0.04), and milk - 0.94 (0.03), while random forest yielded the best predictive performance for egg OFCs with an AUC of 0.96 (0.02). The random forest models were comparable to the LUCCK models for peanut and milk OFC prediction, as the differences in performance were within one standard deviation of each other. In general, the standard deviations of the random forest and LUCCK ensemble models were low across all metrics, indicating the relative stability of these models' performances. The only exception to this was the random forest model for milk OFCs, which had high standard deviation in AUC (0.10) and specificity (0.20).

\begin{figure}[ht]
	\centering
	\includegraphics[scale=.5]{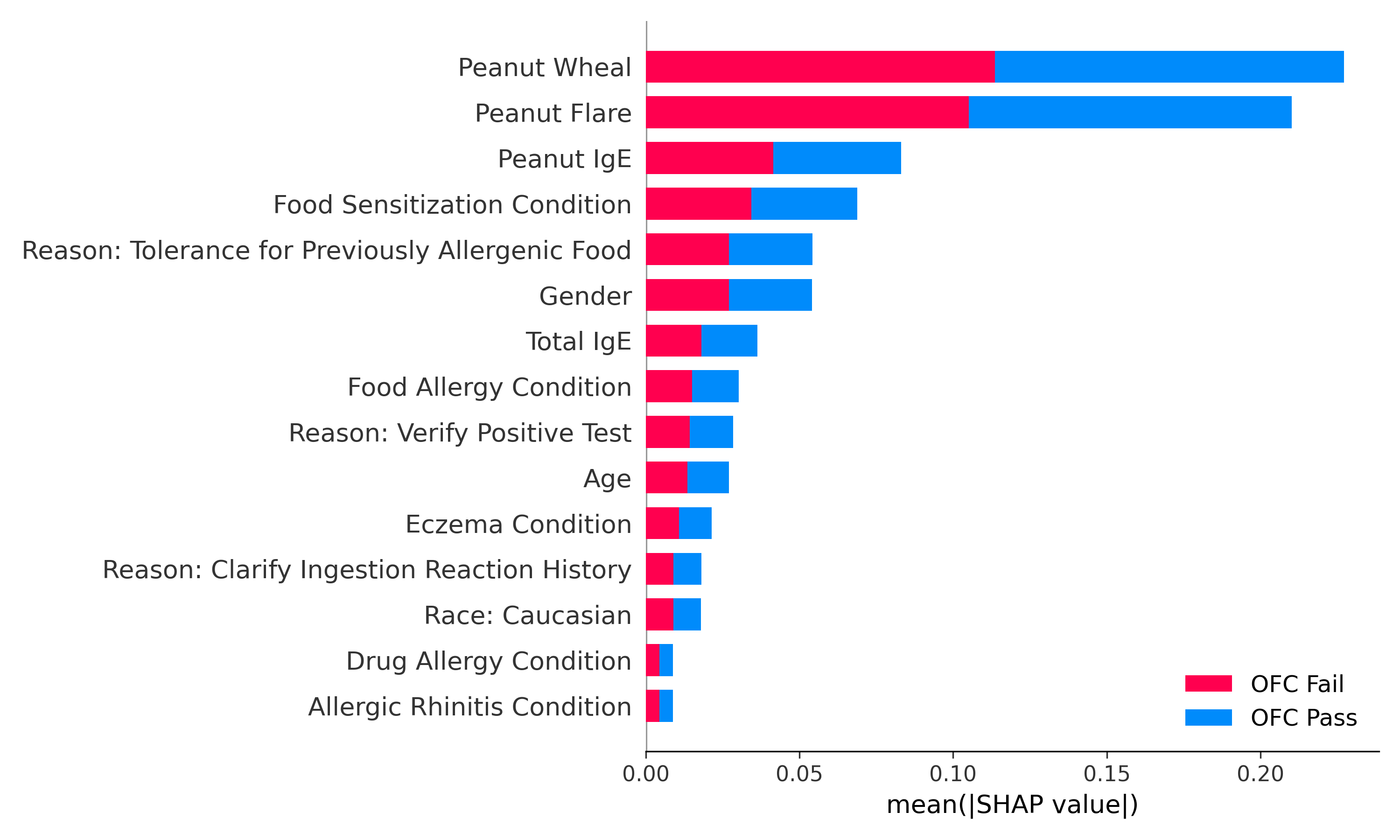}
	\caption{Mean SHAP values for peanut OFCs, coded by class. Larger values correspond to greater predictive contributions.}
	\label{FIG:6}
\end{figure}

\begin{figure}[ht]
	\centering
	\includegraphics[scale=.5]{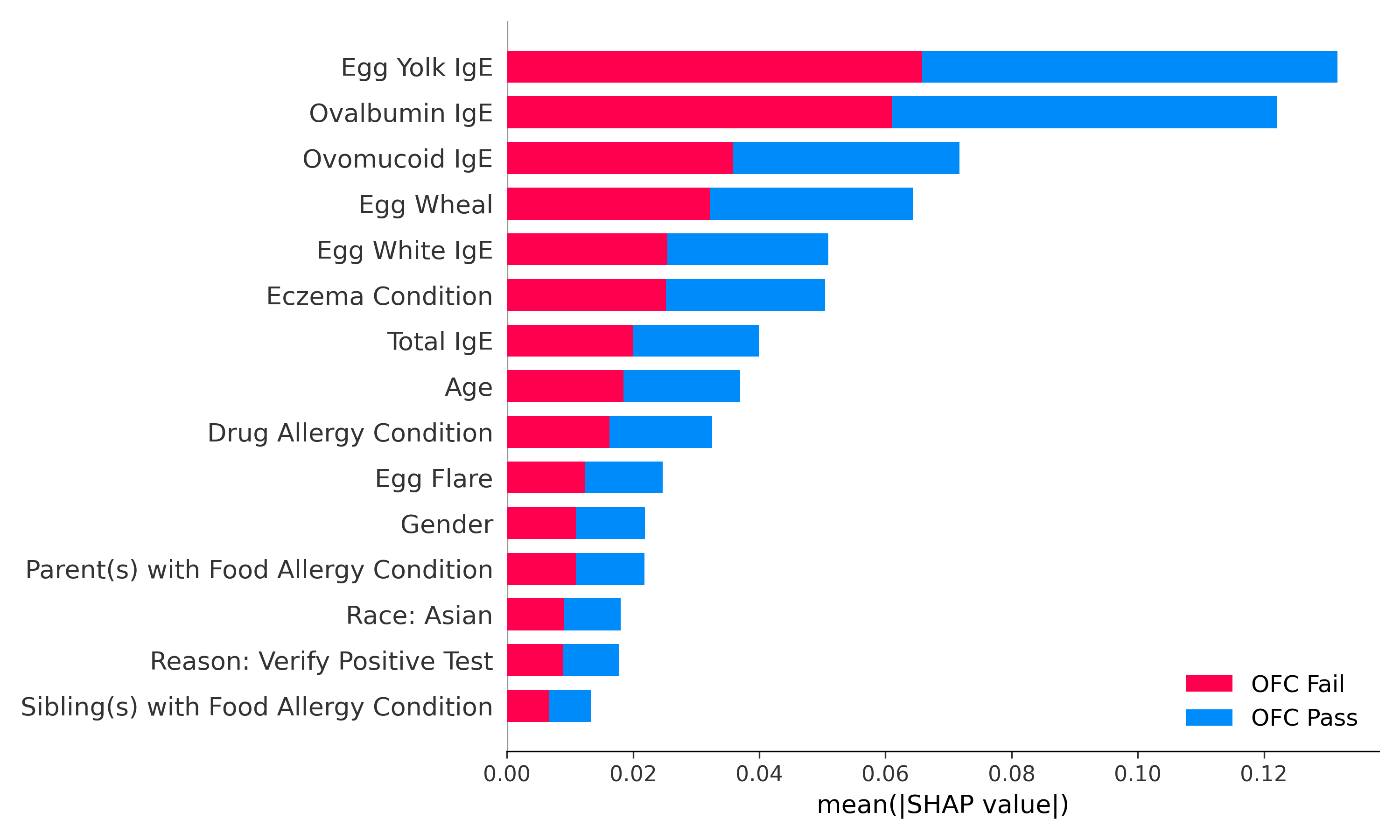}
	\caption{Mean SHAP values for egg OFCs, coded by class. Larger values correspond to greater predictive contributions.}
	\label{FIG:7}
\end{figure}

\begin{figure}[ht]
	\centering
	\includegraphics[scale=.5]{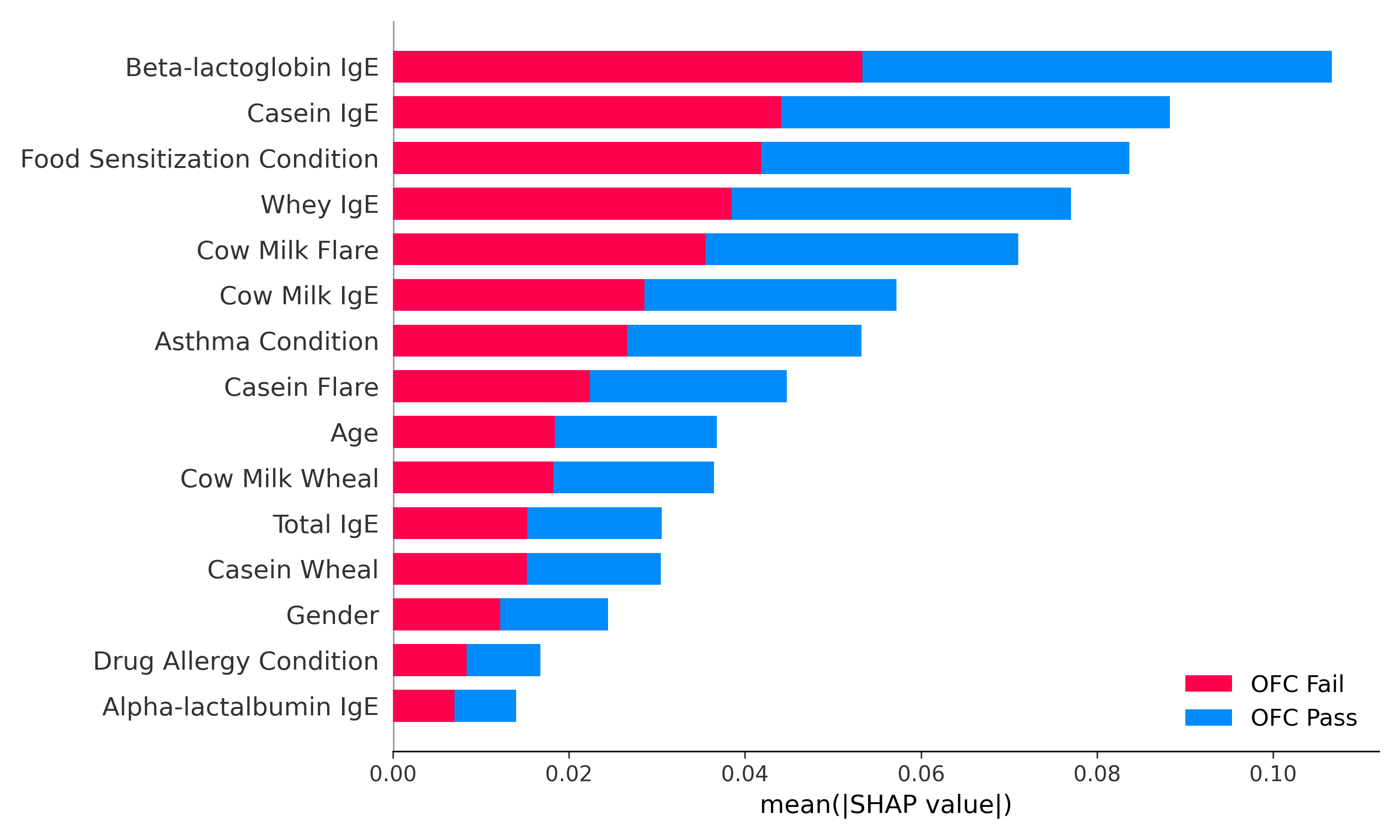}
	\caption{Mean SHAP values for milk OFCs, coded by class. Larger values correspond to greater predictive contributions.}
	\label{FIG:8}
\end{figure}

Model interpretations via SHAP for peanut, egg, and milk OFC predictions are provided in Figures \ref{FIG:6}, \ref{FIG:7}, and \ref{FIG:8}, respectively. The values were derived from the highest performing ensemble model - LUCCK for peanut and milk OFCs, random forest for egg OFCs. In these figures, a higher mean absolute SHAP value indicates that a feature made a greater contribution to the model's OFC prediction, while the proportion of the total SHAP value that corresponds to each class - blue for passed OFCs, red for failed OFCs - indicates the relative importance of each feature in assigning a sample to that class.

Wheal and flare SPT values contributed the most to the prediction of peanut OFC outcomes, with both contributing approximately equally to predictions of both OFC passes and failures. Additional features with high contributions include peanut specific IgE, the patient having food sensitization, and OFCs being administered for the purpose of testing tolerance to a previously known food allergy.  Specific IgE values contributed the most toward egg OFC predictions, in particular egg yolk, ovalbumin, ovomucoid, and egg white, in descending order of contributions. Egg wheal was also found to contribute. For milk OFCs, specific IgE were again among the top five contributors, namely Beta-lactoglobin, casein, and whey, with food sensitization and cow milk flare also contributing. 

\FloatBarrier

\section{Discussion}
\label{sec:discussion}

The best performing ensemble methods proposed in this work achieved accuracy, sensitivity, and specificity values of 90\%, 89\%, and 92\% for peanut OFCs; 98\%, 98\%, and 95\% for egg OFCs, and 94\%, 94\%, and 94\% for milk OFCs. The best performing model for egg OFCs performed significantly better than the model proposed in \cite{pmid33326635}, which achieved accuracy, sensitivity, and specificity values 72\%, 68\%, and 74\%, respectively. In \cite{dunngalvin2011highly}, logistic regression models were created for peanut, egg, and milk OFCs that incorporated six features - age, gender, total and specific IgE, SPT wheal size, and a symptom severity score. The reported accuracy, sensitivity, and specificity of these models were 93\%, 96\%, and 90\% for peanut OFCs; 91\%, 93\%, and 90\% for egg OFCs, and 91\%, 93\%, and 89\% for milk OFCs. These stand in stark contrast to the performance of the equivalent cross-validation logistic regression models in this study:  78\%, 83\%, and 52\% for peanut OFCs; 76\%, 80\%, and 45\% for egg OFCs, and 78\%, 83\%, and 42\% for milk OFCs. This is likely due to the symptom severity score feature utilized in  \cite{dunngalvin2011highly}, which incorporates the severity of a patient's symptoms experienced during \textit{past OFCs}. The incorporation of this feature into a predictive model potentially reduces its utility as it cannot be used without a priori knowledge of past challenge results. In contrast, the proposed models do not presuppose prior knowledge of past OFC outcomes, and primarily rely on IgE, SPT, and comorbidity features as evinced by the SHAP analysis presented in the previous section.

Post-hoc analysis of model performance via SHAP agrees with mRMR predictions in that categorical features contribute less to model predictions than numerical features. For peanut challenges, wheal and flare were found to be the most influential. For egg challenges, both mRMR feature selection and SHAP analysis agreed that ovalbumin IgE and Egg Wheal are important features. For milk, these two disagreed on the importance of whey, and agreed on the importance of the Food Sensitization Condition.

The skin prick test components -- wheal and flare -- were consistently found to be influential in model predictions for all three challenge foods, although significantly more for peanut OFCs. The significance of flare is novel, as flare is typically given low importance by clinicians. In \cite{roberts2005diagnosing}, a peanut wheal of $\ge 8$mm resulted in a likelihood of a failed peanut challenge of 17.25. For our dataset, the highest likelihood for a failed peanut OFC with respect to peanut wheal was 12.06 (95\% CI 11.23-12.89), which occurred when the wheal was $\ge 15$mm, while the highest likelihood for a failed peanut OFC with respect to peanut flare was 4.452 (95\% CI 4.070-4.835), which occurred when the flare was $\ge 31$mm.

The predictive power of flare size demonstrates one of the main advantages in the application of interpretable machine learning to biomedical informatics - the ability to determine heretofore overlooked or underutilized clinical information relevant to the medical phenomena under consideration. Additionally, the improvement in model performance through the use of ensemble methods exemplifies the utility of the probably approximately correct (PAC) learning framework, in which any hypothesis (e.g., a binary function encoding OFC outcomes) that can be strongly learned - a high performance model can be trained using labeled examples - can also be learned by combining weak learners - models trained using labeled examples that perform better than random guessing - to produce a strong one \cite{schapire1990strength}.

The results presented in this study have several limitations. The OFC data utilized in the construction of the machine learning models was collected retrospectively from a single medical center. The proposed models would need to be further validated using data from  multiple external centers to determine their true generalizability. Additionally, the retrospective data was collected over a twenty-one year period, during which changes occurred in the clinical practice of administering OFCs that has likely affected the accuracy and consistency of the collected data. Given that there were an insufficient number of challenges to create separate models for baked vs. raw egg (and similarly for baked vs. raw milk), the relative importance of specific features for those individual allergens may be different than those as determined using the combined models.
 
\FloatBarrier

\section{Conclusion}
\label{sec:conclusion}
In this study, machine learning models were developed to predict OFC outcomes for three common allergens - peanut, milk, and egg. Clinical features such as demographics, comorbidities, IgE levels, and SPT results were obtained from 1,112 patients who collectively underwent a total of 1,284 OFCs. Five machine learning methods - Naive Bayes, Logistic Regression, Support Vector Machines, Random Forest, and Learning Using Concave and Convex Kernels (LUCCK) were evaluated for their predictive performance, either as single models produced via three-fold cross-validation or as an ensemble of three models. The best performing models for all allergens were the random forest and LUCCK ensembles. In terms of area under the curve (AUC), the LUCCK ensemble models achieved the highest performance in OFC prediction for peanut - 0.91, and milk - 0.94, while random forest yielded the best predictive performance for egg OFCs with an AUC of 0.96. Moreover, these models all had sensitivity and specificity values $\geq 89\%$.

Model interpretation via SHapley Additive exPlanations (SHAP) indicated that specific IgE, along with wheal and flare values from SPTs, are highly predictive of OFC outcomes. The significance of flare is novel, as flare is typically given low importance by clinicians in OFC prediction. The results of this analysis suggest that ensemble learning has the potential to predict OFC outcomes with high performance and reveal relevant clinical factors for further study. In future work, other influential features from data modalities such as genomic data and other immunologic measures will be extracted. Additionally, the models will be validated on external datasets from multiple healthcare systems.

\FloatBarrier

\section*{Acknowledgements}
\textbf{Author Contributions:} Conceptualization: JG, DL, RR, GS; Methodology: JZ, JG; Software: JZ, JG; Validation: JG, DL, RR, GS; Formal analysis: JZ;  Data curation: JZ, DS, JG, DL, KJ, JG; Writing - Original Draft: JZ, JG; Writing - Revisions: DL, JG; Supervision: GS, JG; Funding acquisition: JG, KN, RR, GS. All authors have discussed the results and contributed to the final manuscript.

\

\textbf{Funding:} This work was supported by the University of Michigan's Mary H. Weiser Food Allergy Center through its M-FARA Pilot Grant Program.  


\end{document}